\documentclass[review]{elsarticle}
\usepackage{lineno,hyperref}
\usepackage{multirow}
\usepackage{graphicx}
\usepackage{amsmath}
\usepackage{tabularx}
\usepackage{adjustbox}
\usepackage{subfigure}
\usepackage{float}
\usepackage{bm}
\usepackage{hyperref}
\hypersetup{
    colorlinks=true,
    linkcolor=blue,
    filecolor=magenta,      
    urlcolor=cyan,
}

\newcommand{\red}[1]{\textcolor{black}{#1}}
\usepackage{times, graphicx, setspace, color} 
\modulolinenumbers[5]

\usepackage{natbib}
\setcitestyle{authoryear,open={(},close={)}}

\journal{Green Energy and Intelligent Transportation}

\begin{document}

\begin{frontmatter}

\title{Deep Transfer Learning for Intelligent Vehicle Perception: a Survey}

\author{Xinyu Liu$^{1}$, Jinlong Li$^{1}$, Jin Ma$^{1}$, Huiming Sun$^{1}$, Zhigang Xu$^{2}$, \\ 
Tianyun Zhang$^1$, Hongkai Yu$^1$*}
\address{
  $^1$ Department of Electrical Engineering and Computer Science, Cleveland State University, Cleveland, OH 44115, USA 
  \\
  $^2$ School of Information Engineering, Chang’an University, Xi’an 710064, China
  }


\cortext[Hongkai Yu]{Corresponding author: Hongkai Yu, Email: h.yu19@csuohio.edu.}

\begin{abstract}
Deep learning-based intelligent vehicle perception has been developing prominently in recent years to provide a reliable source for motion planning and decision making in autonomous driving. A large number of powerful deep learning-based methods  can achieve excellent performance in solving various perception problems of autonomous driving. However, these deep learning methods still have several limitations, for example, the assumption that lab-training (source domain) and real-testing (target domain) data follow the same feature distribution may not be practical in the real world. There is often a dramatic  domain gap between them in many real-world cases. As a solution to this challenge, deep transfer learning can handle situations excellently by transferring the knowledge from one domain to another. Deep transfer learning aims to improve task performance in a new domain by leveraging the knowledge of similar tasks learned in another domain before. Nevertheless, there are currently no survey papers on the topic of deep transfer learning for intelligent vehicle perception. To the best of our knowledge, this paper represents the first comprehensive survey on the topic of the deep transfer learning for intelligent vehicle perception. This paper discusses the domain gaps related to the differences of sensor, data, and model for the intelligent vehicle perception. The recent applications, challenges, future researches in intelligent vehicle perception are also explored.  

\end{abstract}
 

\end{frontmatter}


\section{Introduction}\label{introduction}

In recent years, perception has been viewed as a critical component in intelligent vehicles for precise localization, safe motion planning, and robust control~\cite{li2020intelligent, yurtsever2020survey,huang2020autonomous}. The perception system provides intelligent vehicles with immediate environmental information about surrounding pedestrians, vehicles, traffic signs, and other items and helps to avoid possible collisions. \textcolor{black}{In this paper, the terminology ``perception" is mainly focused on the detection and segmentation tasks by ignoring the tracking and trajectory prediction tasks. This focus is because of  two reasons: (1) Some previous intelligent vehicle research works mainly use detection and/or segmentation tasks to describe the terminology ``perception" for intelligent vehicles~\cite{van2018autonomous,xu2022opv2v}; (2) Many downstream tasks, such as tracking, trajectory prediction, and behavior prediction, are dependent on the accurate detection and/or segmentation first.}

\begin{figure*}[htbp]
    \begin{centering}
        \includegraphics[width=0.9\textwidth]{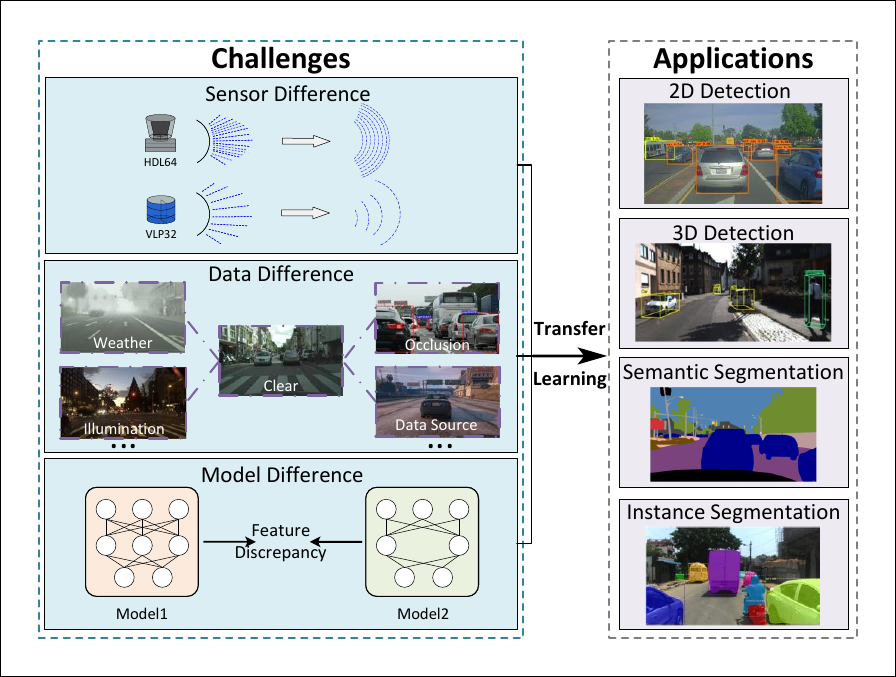}
        \par\end{centering}
    \caption{\textcolor{black}{Illustration of Challenges and Applications of Intelligent Vehicle Perception. Transfer Learning (TL)  methods can be applied to reduce the domain gaps by sensor difference, data difference, and model difference.}}
    \label{fig:definition}
\end{figure*}

The perception tasks play an indispensable role in intelligent vehicles and autonomous driving~\cite{arnold2019survey}. Recently, the deep learning methods have gained significant traction in the intelligent vehicle perception and have achieved great successes~\cite{grigorescu2020survey,wen2022deep,chen2022milestones}. \textcolor{black}{For example, as shown in Fig.~\ref{fig:definition}, four important applications have been actively studied and have gained significant advancements in recent years. (1) 2D object detection is a prominent task which aims at recognizing and localizing objects within the images or videos. The main goal is to develop algorithms or models that can automatically detect objects and accurately outline their boundaries. (2) 3D object detection is a crucial task that focuses on discerning and precisely localizing objects within a three-dimensional space. 3D detection includes estimating the position of an object and orientation within a comprehensive 3D coordinate system. (3) Semantic segmentation is a pixel-level image analysis task where each pixel in an image is assigned a semantic label. The goal is to partition the image into coherent regions or segments based on the objects or classes they belong to. (4) Instance segmentation combines the object detection and semantic segmentation task. The objective is to detect and segment individual objects within an image, providing a unique label and pixel-level mask to each instance.}


\textcolor{black}{However, as shown in Fig.~\ref{fig:definition}, there are lots of complex cases where the deep learning methods might fail in the real world. For example, a deep learning based vehicle detection model pre-trained on clear weather data might be then tested in the foggy weather, low-illumination (like night), occlusion, or different  data source (like simulation) conditions, leading to a large performance drop}. This degradation is influenced by the domain gap (shift) between diverse driving  environments~\cite{hnewa2020object,mirza2022efficient,mohammed2020perception}, \textit{e.g.}, \textcolor{black}{different  weather, illumination, occlusion, data source conditions}. Moreover, different types and settings of the sensors~\cite{rist2019cross} installed on vehicles and various deep learning model structures~\cite{xu2023bridging}~\cite{khalil2022licanet} during the Vehicle-to-Vehicle (V2V) cooperative perception might result in the domain gap as well. 

The above mentioned performance drop for intelligent vehicle perception because of the domain gap can be relieved via the Transfer Learning (TL)~\cite{zhuang2020comprehensive} methods. The TL techniques include two goals: (1) fully using the prior knowledge obtained from the source domain to guide the inference in the related target domain, (2) largely reducing the feature distribution discrepancy caused by the domain gap. Due to these two goals, the performance for deep learning based intelligent vehicle perception systems in related but different domains can be enhanced. The deep learning model's generalization capability can be improved for the intelligent vehicle perception under different challenging scenarios as shown in Fig.~\ref{fig:definition}.

In this paper, we focus on the transfer learning methods for the intelligent vehicle perception in the deep learning era. This paper first reviews the related tasks and benchmark datasets for intelligent vehicle perception, and then classifies the domain gaps to three differences of sensor, data, and model during the vehicle driving. Next, we carefully review about 150 related published papers of deep transfer learning since the deep learning research is started, then we classify the deep learning based transfer learning methods into four types: (1) Supervised TL, (2) Unsupervised TL, (3) Weakly-and-semi Supervised TL, and (4) Domain Generalization. For the first three types, the transfer learning is implemented from one source domain to one target domain, where our classification depends on whether the target domain has labeled data or not. For the last type, the transfer learning is conducted from one source domain to multiple target domains for the generalization in many seen or unseen driving scenarios. In addition, several subdivisions of each type of transfer learning methods are also reviewed and analyzed in this survey.

The contributions of this paper can be outlined as follows. 

\begin{itemize}
    \item To the best of our knowledge, this paper is the first in-depth survey on the topic of the deep transfer learning for intelligent vehicle perception. 
    
    \item This paper summarizes the domain gap for intelligent vehicle perception into three types (differences of sensor, data, model) and gives detailed explanations to the related tasks and benchmark datasets. 

    \item After reviewing about 150 related published papers, we classify the deep transfer learning methods for intelligent vehicle perception into four types and explain each of them in details.    
\end{itemize}

The subsequent sections of this paper are structured as follows. Section~\ref{sec:iv-p} provides a general overview about the related tasks and benchmark datasets for intelligent vehicle perception. Section~\ref{sec:DD-D} presents the domain distribution discrepancy and the three kinds of domain gap. Section~\ref{sec:TL-M} details the different methodologies of the deep transfer learning techniques. Section~\ref{sec:Cha} explains the challenges and future research, followed by a conclusion in Section~\ref{sec:Con}.

\section{Intelligent Vehicle Perception}\label{sec:iv-p}

For intelligent vehicles or autonomous driving, perception plays a crucial role in receiving data from sensors and extracting meaningful information from the surrounding environment, so as to make meaningful decisions for the precise motion planning by identifying  obstacles, traffic signs/markers, and available driving areas~\cite{li2023continual}. Two  types of mainstream sensors (Camera, LiDAR) are widely used in self-driving or intelligent driving vehicles~\cite{cao2019adversarial}~\cite{fadadu2022multi}~\cite{liu2021automated}~\cite{liu2022yolov5}~\cite{gholamhosseinian2021vehicle, yu2022review}. These sensors installed on vehicles are utilized for the intelligent vehicle perception tasks. 

The intelligent vehicle perception tasks include discovering the surrounding vehicles and pedestrians, recognizing traffic signs and markers, finding the driving areas (\textit{e.g.}, road regions), and so on. In the real world, sometimes the objects may be similar to each other or the background, and the challenging scenarios (\textit{e.g.}, diverse weather, dark illumination) might affect the performance of sensors, making the perception tasks even more difficult~\cite{hnewa2020object,li2023domain}. This paper groups these intelligent vehicle perception tasks into two classes (Object Detection, Semantic/Instance Segmentation) and further discusses these challenges for intelligent vehicle perception in the real world.

\subsection{Object Detection} 

To achieve autonomous driving safely and  successfully, it is necessary to have a reliable object detection system. Considering the complex road conditions, it is essential to detect (localize and recognize) other vehicles, pedestrians, and obstacles to prevent potential accidents. However, detecting objects in urban areas is challenging due to the diverse types of objects and unknown road situations~\cite{arnold2019survey,feng2020deep}.

\textbf{2D Object Detection:} By only using the relatively cheap camera sensor(s), deep learning models can be easily applied to efficiently detect (localize and recognize) the surrounding objects from the 2D image data~\cite{yeong2021sensor}. The output will be the identified 2D bounding boxes (2D coordinates) with the recognized object classes for the surrounding objects on each camera image, with a real-time or near real-time inference speed. However, 2D object detection alone can only provide the object's position on a 2D plane, which does not provide enough information~\cite{wang2019pseudo}, \textit{e.g.}, object depth, object 3D size.

\textbf{3D Object Detection:} Considering the limitations of 2D object detection, the object 3D information might equip the intelligent vehicle with the capability to more robustly and accurately perceive and recognize surrounding objects. The output will be the identified 3D bounding boxes (3D coordinates) with the recognized object classes for the surrounding objects, with a reasonable inference time. 
Because the images of camera sensors and the point clouds of LiDAR sensors could provide the depth cues, the 3D object detection task could be achieved via three sensor settings: (1) Camera only~\cite{wang2023multi}, (2) LiDAR only~\cite{xu2023v2v4real}~\cite{xu2022opv2v}~\cite{li2023s2r}, (3) Camera + LiDAR~\cite{zhao2020fusion}.

\subsection{Semantic/Instance Segmentation}
Different with the object detection task, the segmentation task not only discovers the object regions but also give the pixel-level labels (masks) for everything (object and background) in the driving scenarios. For the intelligent vehicle perception, the segmentation task can be classified into two types: Semantic Segmentation, Instance Segmentation. 

\textbf{Semantic Segmentation:} Semantic segmentation involves the assignment of a semantic label to every pixel within an image, such as ``road", ``vehicle" or ``pedestrian", ``traffic sign", and so on. This technique enables the intelligent vehicle to perceive the surrounding environment and understand the scene more comprehensively~\cite{feng2020deep,mo2022review}. The identification of specific regions within an image can aid the self-driving vehicles in making informed decisions, \textit{e.g.}, determining where the driving road region is.

\textbf{Instance Segmentation:} Instance segmentation outputs the boundaries (pixel-level masks) of each object and assigns a unique label to each discovered object~\cite{zhou2020joint}, which seems like a integration of object detection and semantic segmentation. It is particularly useful for identifying the shape, location, and number of surrounding objects in autonomous driving~\cite{rashed2021generalized,ko2021key}.

\textcolor{black}{Semantic segmentation captures the overall scene structure, while instance segmentation enables a more fine-grained understanding of objects and their boundaries. Different from semantic segmentation which only classifies images into meaningful semantic regions, instance segmentation provides more precise analysis by supplying a separate semantic mask (with identity) for every object instance.}

\subsection{Benchmark Dataset}
\textcolor{black}{There might be many different sensors installed on the intelligent vehicles, such as  Camera, LiDAR, Radar, Near-Infrared Sensors, Ultrasonic Sensors, and so on. For a clear description, we focus on introducing the two widely-used main sensors (Camera and LiDAR) in this paper, related to the image data by Camera and the point cloud data by LiDAR on the intelligent vehicles.}

\textbf{Camera data:} The 3-channel color images in Green, Red, Blue primary colors of light (\textit{i.e.}, RGB images)   are commonly acquired by monocular or multiple cameras, which are simple and reliable sensors that closely resemble human eyes~\cite{feng2020deep}. One of the main benefits of RGB cameras is their high resolution and relatively low cost. However, their performance can deteriorate significantly under the challenging weather and illumination conditions~\cite{feng2021review}. 
 
\textbf{LiDAR data:} Unlike cameras, laser sensors offer direct and precise 3D information, making it easier to extract object candidates and aiding in the classification task by providing 3D shape information. LiDAR, also known as light detection and ranging, is a sensor technology that is capable of detecting targets in all lighting conditions and creating a distance map of the targets with high spatial coverage~\cite{li2020lidar,li2020deep}. LiDAR could work in some challenging weather and dark illumination scenarios, but it is quite expensive with high cost. Its high cost is a major obstacle to wider adoption~\cite{li2020deep}~\cite{pham20203d}.

\textbf{Benchmark for 2D Object Detection:} KITTI~\cite{geiger2013vision}, Cityscapes~\cite{cordts2016cityscapes}, SIM10k~\cite{johnson2017driving}, Foggy  Cityscapes~\cite{sakaridis2018semantic}, Syn2Real-D~\cite{peng2018syn2real},  BDD100k~\cite{yu2018bdd100k},  GTA5~\cite{richter2016playing},  nuScenes~\cite{caesar2020nuscenes}, 
Waymo Open~\cite{sun2020scalability},  A*3D~\cite{pham20203d},  ApolloScape~\cite{huang2018apolloscape}, Ford~\cite{agarwal2020ford},  A2D2~\cite{geyer2020a2d2}, 
ONCE~\cite{mao2021one}, 
and Automine~\cite{li2022automine}. \textcolor{black}{Let us give two examples:} (1) \textcolor{black}{KITTI~\cite{geiger2013vision} is a widely utilized dataset for autonomous driving, encompassing camera images, LiDAR point clouds, and ground-truth 3D bounding boxes. It comprises 7,481 training frames and 7,518 test frames, accompanied by sensor calibration data and annotated 3D bounding boxes around objects of interest} (2) \textcolor{black}{SIM10k~\cite{johnson2017driving} is a synthetic dataset derived from the Grand Theft Auto V (GTA-V) computer game. It contains 10,000 images capturing driving street scenes, accompanied by bounding box annotations specifically for cars.}

\textbf{Benchmark for 3D Object Detection:}  KITTI~\cite{geiger2013vision}, Cityscapes~\cite{cordts2016cityscapes}, 
Foggy Cityscapes~\cite{sakaridis2018semantic}, 
GTA5-LiDAR~\cite{wu2019squeezesegv2},  nuScenes~\cite{caesar2020nuscenes}, 
Waymo Open~\cite{sun2020scalability},  A*3D~\cite{pham20203d},  ApolloScape~\cite{huang2018apolloscape},  Ford~\cite{agarwal2020ford},  A2D2~\cite{geyer2020a2d2}, 
ONCE~\cite{mao2021one}, 
Automine~\cite{li2022automine}, OPV2V~\cite{xu2022opv2v} and V2V4Real~\cite{xu2023v2v4real}. \textcolor{black}{Here we explain two examples:} (1) \textcolor{black}{The Waymo Open Dataset~\cite{sun2020scalability} is a Camera+LiDAR dataset, which consists of 1,000 driving sequences containing 798 scenes allocated for training and 202 scenes allocated for validation. The evaluation metrics employed are Average Precision(AP) and Average Precision with Heading (APH) information, which consider the weighted average precision and the heading accuracy, respectively. The metrics are computed based on the 3D Intersection Over Union (IoU) threshold of 0.7 for vehicles and 0.5 for others.} (2) \textcolor{black}{V2V4Real dataset~\cite{xu2023v2v4real} is designed for the connected vehicles based cooperative perception of autonomous driving using V2V (Vehicle-to-Vehicle) communication, encompassing a driving area spanning 410 km. It includes a substantial collection of real-world data, such as 20,000 LiDAR frames, 40,000 RGB frames, 240,000 annotated 3D bounding boxes for 5 specific classes, and comprehensive High-Definition (HD) Maps containing all driving routes}.

\textbf{Benchmark for Semantic Segmentation:} KITTI~\cite{geiger2013vision}, Cityscapes~\cite{cordts2016cityscapes}, ApolloScape~\cite{huang2018apolloscape}, BDD100k~\cite{yu2018bdd100k}, and A2D2~\cite{geyer2020a2d2}. \textcolor{black}{Two examples are given here:} (1) \textcolor{black}{Cityscapes~\cite{cordts2016cityscapes} is a semantic urban scene dataset under driving scenarios. It includes semantic and instance segmentation annotations. The dataset comprises 2,975 training images with a resolution of 2048 × 1024, along with an additional set of 500 validation images.} (2) \textcolor{black}{ApolloScape~\cite{huang2018apolloscape} contains a large collection of sequentially recorded 140,000 camera images with pixel-level semantic annotations in different driving conditions. It comprises 40,960 training images and 8,327 validation images. In addition to the semantic annotations, the dataset also includes pose information relative to static background point clouds.}

\textbf{Benchmark for Instance Segmentation:} Cityscapes~\cite{cordts2016cityscapes},  nuScenes~\cite{caesar2020nuscenes}, 
BDD100k~\cite{yu2018bdd100k}, 
and KITTI-360~\cite{liao2022kitti}. \textcolor{black}{Let us give two examples:} (1) \textcolor{black}{BDD100k~\cite{yu2018bdd100k} includes 100,000 images with a resolution of 1280 × 720, and it has training, testing and validation sets. Among them, annotated 70,000 images are for training and annotated 10,000 images are designed for validation. This dataset contains six various weather conditions, six distinct scenes, three different parts of a day, and ten object categories with bounding box annotations. (2) nuScenes~\cite{caesar2020nuscenes} is an autonomous driving dataset comprising 1,000 driving scenes. In addition to the scene annotations, nuScenes also provides High-Definition (HD) semantic maps, offering insights into 11 distinct semantic classes. This dataset encompasses 700 scenes for training, 150 scenes for validation, and another 150 scenes for testing. The data collection process involved the utilization of six cameras and a 32-beam LiDAR system, while the annotations cover 10 objects within a complete 360-degree field of view.}

The Table~\ref{tab:Benchmark} summarizes the current widely-used benchmark dataset details for the intelligent vehicle perception tasks, including the image resolution, image numbers, LiDAR frame numbers, task types, real or synthetic information of each benchmark dataset.

\begin{table}[]
    \centering
    \footnotesize
\caption{\label{tab:Benchmark} Benchmark Datasets for Intelligent Vehicle Perception. D: object detection in 2D or 3D, S: semantic or instance segmentation, Syn: synthetic data, R: real data.}
        \begin{adjustbox}{width=1\textwidth}
        \begin{tabular}{c|c|c|c|c|c}
            \hline
            \hline
            \textbf{Benchmark}   & Image Resolution    & Image \#   & LiDAR Frame \#    & Tasks & Real/Syn \tabularnewline \hline
            KITTI~\cite{geiger2013vision}    & 1,392×512   & 15K & 1.3M  & D, S & R\tabularnewline \hline
            Cityscapes~\cite{cordts2016cityscapes}    & 2,048×1,024  & 25K & - & D, S & R \tabularnewline \hline
            SIM10k~\cite{johnson2017driving}    & 1,914×1,052   & 10K & - & D & Syn\tabularnewline \hline
            Foggy Cityscapes~\cite{sakaridis2018semantic}    & 2,048×1,024   & 3,475 &  - & D, S & Syn \tabularnewline \hline
            Syn2Real-D~\cite{peng2018syn2real}    &  -  & 248K  & - & D & Syn, R\tabularnewline \hline
            BDD100K~\cite{yu2018bdd100k}    & 1,280×720   & 8K  & - & D, S & R\tabularnewline \hline
            GTA~\cite{richter2016playing}    & 1,914×1,052   & 24,966 & - & S  & Syn\tabularnewline \hline
            GTA-LiDAR~\cite{wu2019squeezesegv2}    & 64×512   & 100K & - & S & Syn\tabularnewline \hline
            H3D~\cite{patil2019h3d}    & 1,920×1,200   & 27,721 & - & D  & R\tabularnewline \hline
            nuScenes~\cite{caesar2020nuscenes}    & 1,600×900   & 40K &  -  & D & R\tabularnewline \hline
            Waymo Open~\cite{sun2020scalability}    & 1,920×1,280   & 200K & - & D & R\tabularnewline \hline
            ApolloCar3D~\cite{song2019apollocar3d}    & 3,384×2,710 &  5,277 & - & S & R\tabularnewline \hline
            A*3D~\cite{pham20203d}    & 2,048×1,536  &  39K & 39,179 & D & R\tabularnewline \hline
            ApolloScape~\cite{huang2018apolloscape}    & 3,384×2,710 &   143,906  & - & S & R\tabularnewline \hline
            SYNTHIA~\cite{ros2016synthia}    & 960×720  & 13.4K & - & S & Syn\tabularnewline \hline
            Lyft Level 5~\cite{houston2021one}    & -  & 55K & - & S & R\tabularnewline \hline
            Ford~\cite{agarwal2020ford}    & -  & 200K & - & D & R\tabularnewline \hline
            A2D2~\cite{geyer2020a2d2}    &  1,928×1,208  & 12K &   & D, S & R \tabularnewline \hline
            ONCE~\cite{mao2021one}    &  1,920×1,020  &  1M & - & D & R \tabularnewline \hline
            AutoMine~\cite{li2022automine}    &  2,048×1,536  &  18K & - & D & R \tabularnewline \hline
            OPV2V~\cite{xu2022opv2v}    &  800×600  &  44K & 11K & D & Syn \tabularnewline \hline
            V2V4Real~\cite{xu2023v2v4real}    &  2,064×1,544  & 40K & 20K & D & R \tabularnewline \hline
            \hline        		
        \end{tabular}
        \end{adjustbox}
    
\end{table}

\begin{figure*}[htbp]
    \begin{centering}
        \includegraphics[width=1\textwidth]{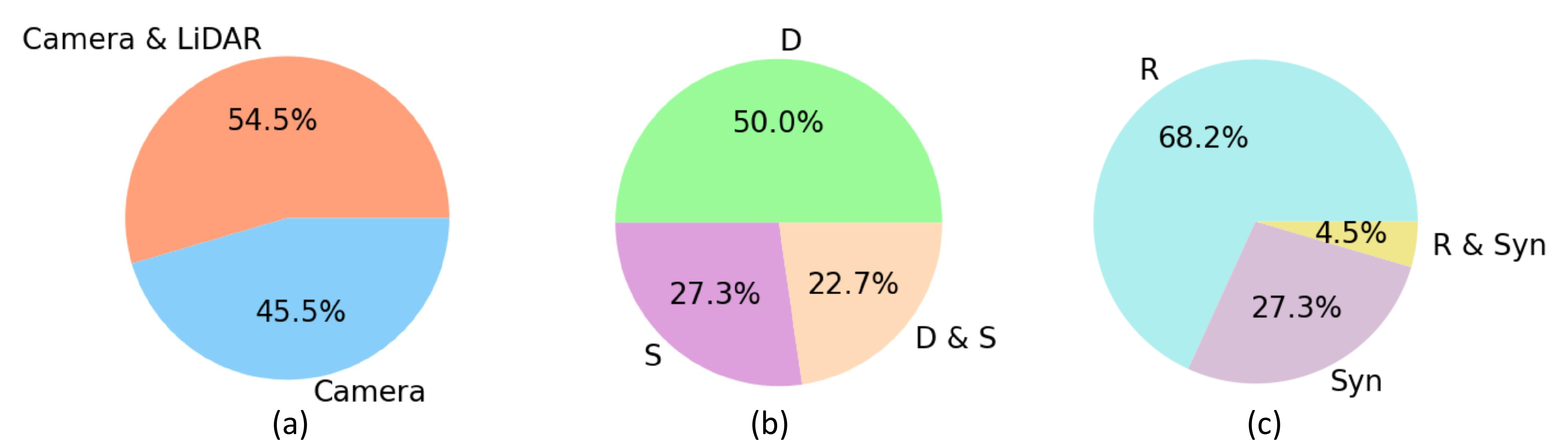}
        \par\end{centering}
    \caption{\textcolor{black}{Distribution Pie Charts of the Benchmark Datasets for Intelligent Vehicle Perception: (a) sensor type, (b) task, (c) data source. Note: D: object detection in 2D or 3D, S: semantic or instance segmentation, Syn: synthetic data, R: real data.}}
    \label{fig:pie}
\end{figure*}

\section{Domain Distribution Discrepancy}\label{sec:DD-D}
Despite the remarkable achievements of the intelligent vehicle perception  algorithms on benchmark datasets, there are still significant challenges in the real world due to the large variations in the sensor types and settings, data in diverse style, environment, weather and illumination, trained epoch, and  architecture~\cite{li2022cross,feng2021review,schutera2020night,song2023synthetic}. Based on these observations, we divide the domain distribution discrepancy for intelligent vehicle perception into three types: sensor difference, data difference, and model difference, as shown in Table~\ref{Domain Distribution Discrepancy}.

\subsection{Sensor Difference}
First of all, \textcolor{black}{the domain gap shows up when the sensors are different in types and settings as described  in~\cite{chen2023milestones, wang2023multi, li2022survey}}. Let us explain the sensor difference for camera and LiDAR separately. The camera sensor is cheap but not robust to different types and settings, for example, angle difference from horizontal to oblique~\cite{rist2019cross}, placement dissimilarity from front view to rear view~\cite{alonso2020domain}, image resolution diversity~\cite{carranza2020performance}, and so on. The LiDAR sensors might also have different types and settings, for example, different laser  beam numbers~\cite{yi2021complete}, various LiDAR equipment from different companies~\cite{xu2023bridging}, LiDAR placement dissimilarity~\cite{hu2022investigating}, and so on. \textcolor{black}{The problem is quite similar to the setting of other sensor types. Taking Radar as an example, variations in Radar resolutions, field-of-view, and noise characteristics can result in diverse data distributions.} These real-world challenges due to the sensor difference may generate the heterogeneous feature distribution between different domains~\cite{triess2021survey,zhou2022towards,chakeri2021platform}.

\subsection{Data Difference}
In addition, \textcolor{black}{the domain gap exists when the data itself is different in style and format as explained in~\cite{gao2021we, triess2021survey}. The data collected by the sensors in different situations might result in the heterogeneous data distribution between different domains. See the following four common examples for intelligent vehicles. (1) Diverse Weather: foggy, rainy, snowy, sunny,  \textit{etc}~\cite{miglani2019deep,xu2021spg,mirza2022efficient,bogdoll2022anomaly,li2023domain}. (2) Various Illumination: daytime, nighttime, tunnel, \textit{etc}~\cite{wu2021dannet}. (3) Occlusion: objects or parts of objects are obscured or hidden from the sensors' view due to obstacles, other vehicles, or environmental conditions~\cite{qian20223d, ruan2023review}. (4) Different Data Source: differences between synthetic data (computer game data: SYNTHIA~\cite{ros2016synthia}, GTA5~\cite{richter2016playing}) and real-world data~\cite{li2023s2r}, and the changes between the data collected in different urban or highway environments~\cite{shenaj2023learning}.}

\subsection{Model Difference}
Finally, \textcolor{black}{the model difference is not common but it is also one possible reason for the domain gap in intelligent vehicle perception}. When the perception architecture is diverse, the model is obviously different~\cite{xu2023bridging}, for example, from PointPillar architecture~\cite{lang2019pointpillars} to SECOND architecture~\cite{yan2018second}. When the perception architecture is the same, the domain gap may still exist because of heterogeneous configurations, like training epochs, resolution, number of convolution layers, and hyperparameters~\cite{xu2023bridging}. Because of these diverse situations, the features extracted from them might have a domain shift, as described in~\cite{xu2023bridging}, leading to the heterogeneous feature distribution between different domains.

\begin{table}[h]
\centering
\footnotesize

\caption{\red{Domain Distribution Discrepancy with Three Types of Differences for Intelligent Vehicle Perception: sensor, data and model. ``$\rightarrow$" means the model training with the left data and testing on the right data.}}
\label{Domain Distribution Discrepancy}

\begin{adjustbox}{width=1\textwidth}
\begin{tabular}{c|c|c}
\hline
\hline
\textbf{Types} & \textbf{Differences} & \textbf{Examples} \\
\hline
\multirow{3}{*}{Sensor Difference} & Setup & 64-beam LiDAR $\rightarrow$ 32-beam LiDAR~\cite{yi2021complete}\\
& Placement & Front $\rightarrow$ Rear~\cite{feng2020deep}\\
& Angle & Horizontal $\rightarrow$ Oblique~\cite{rist2019cross}\\
\hline
\multirow{3}{*}{Data Difference} & \textcolor{black}{Data Source} & GTA5 $\rightarrow$ Cityscapes~\cite{murez2018image}\\
& \textcolor{black}{Occlusion} & OPV2V $\rightarrow$ V2V4Real~\cite{xu2023v2v4real}\\
& Weather & Cityscapes $\rightarrow$ Foggy Cityscapes~\cite{li2023domain}\\
& Illumination & Cityscapes $\rightarrow$ Dark Zurich~\cite{wu2021dannet}\\
\hline
\multirow{3}{*}{Model Difference} & \textcolor{black}{Configuration} & Hyperparameter 1 $\rightarrow$ Hyperparameter 2 ~\cite{xu2023bridging}\\
& \textcolor{black}{Architecture} & PointPillars $\rightarrow$ SECOND~\cite{xu2023bridging}\\
\hline
\hline
    \end{tabular}
    \end{adjustbox}

\end{table}

\section{Deep Transfer Learning Methodology} \label{sec:TL-M}
With the rapid advancement of autonomous driving techniques, there is now an abundance of driving scene images available. Deep learning methods are booming in the application of autonomous driving with high  performance of perception. This paper is focused on the transfer learning methods for the intelligent vehicle perception in the deep learning era.

\begin{figure*}[]
    \begin{centering}
        \includegraphics[width=1\textwidth]{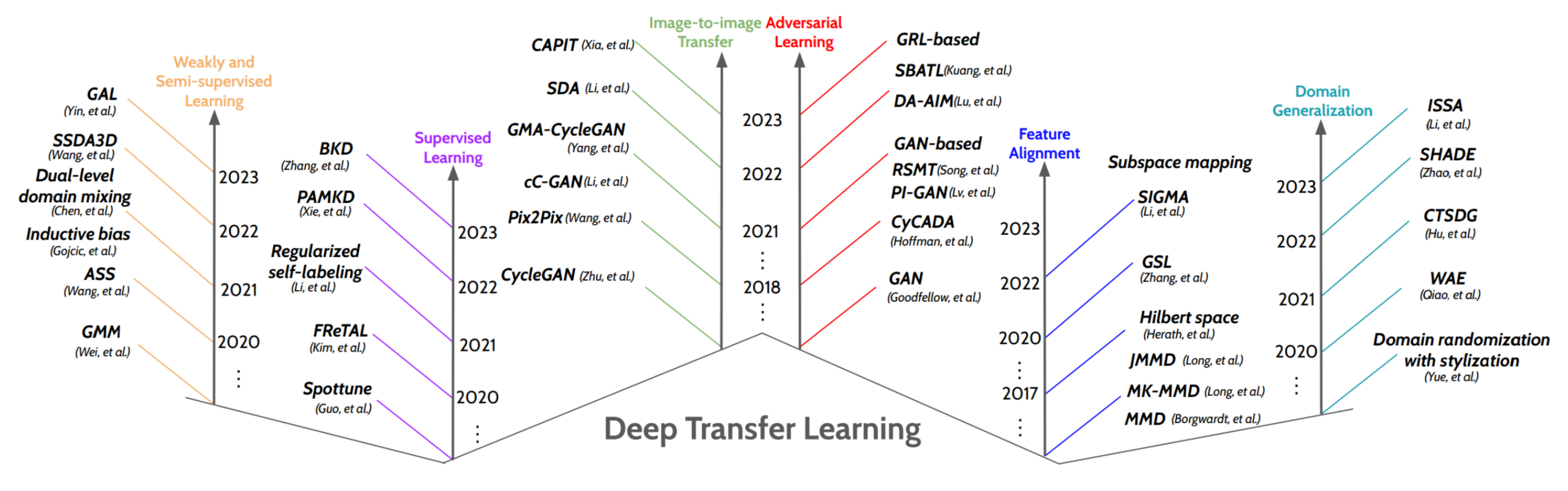}
        \par\end{centering}
    \caption{\textcolor{black}{The Chronological Overview of Transfer Learning Research in the Deep Learning Era}.}
    \label{fig:overview}
\end{figure*}

Transfer Learning (TL) is a machine learning method to largely apply the knowledge acquired from one task or domain to another related task or domain~\cite{zhuang2020comprehensive}. This paper classifies the deep transfer learning into several main types: Supervised TL, Unsupervised TL, Weakly-and-semi Supervised TL, Domain Generalization. The chronological overview of the transfer learning research development in the deep learning era is shown in Fig.~\ref{fig:overview}.

\subsection{Supervised TL}
In the transfer learning research, the source domain normally has the manually annotated ground truth. If the target domain also has the manually annotated ground truth, the machine learning technique that transfers knowledge from a labeled source domain to the labeled target domain is named as Supervised TL~\cite{drews2017aggressive,yu2018bdd100k,zhou2019autonomous}. Gathering such manually annotated data requires substantial human involvement, which is labor-intensive and time-consuming~\cite{carvalho2015automated}. 

We divide the Supervise TL methods into Fine-tuning and Knowledge distillation via teacher-student network in this paper.

\subsubsection{Fine-tuning}

Fine-tuning is a common technique in the transfer learning~\cite{guo2019spottune,li2021improved,hu2022investigating}, which has been widely used in intelligent vehicle perception~\cite{wang2019pseudo,luo2021self,liang2022federated,xu2019training,doan2019scalable}. Fine-tuning takes an existing neural network model pre-trained on a source domain dataset and further trains it on a new target domain dataset. By the fine-tuning, the knowledge learned from the source domain can be leveraged to improve the performance on the target domain. It is worth mentioning that fine-tuning a pre-trained neural network model could obtain better performance than directly training from scratch. Typically, the pre-trained neural network model is trained on a large-scale dataset, enabling to acquire the knowledge from a wide range. The learning rate of fine-tuning on the target domain is relatively small as a fine adjustment for the neural network model pre-trained on source domain. The fine-tuning methods could be roughly classified into two types: (1) \textit{Whole Fine-tuning}: it trains all the layers of the whole neural network model. (2) \textit{Partial Fine-tuning}: it allows us to train only the some interested layers of the pre-trained neural network while keeping the some layers frozen.      

\textit{Whole Fine-tuning}: All the layers of the entire neural network model are fine-tuned to obtain the spatial–temporal interactions~\cite{ye2021gsan} among autonomous vehicles and the 3D perception in autonomous driving~\cite{sautier2022image}. 

\textit{Partial Fine-tuning}: \cite{guo2018small} only fine-tunes the encoder-decoder based semantic segmentation model, by fixing a pre-trained sub-network to ensure the multi-class boundary constraint.

\textcolor{black}{\textbf{In-depth Discussion:} (1) The fine-tuning based transfer learning is a simple but effective way to transfer the knowledge gained from the pre-training on source domain to enhance the performance on the target domain with less data and computational resources than training from scratch. The training iterations are typically fewer when we fine-tune a pre-trained model, compared to training a machine learning model from scratch. Because the pre-trained model already has the prior knowledge of the pre-training datasets (normally large), the fine-tuning process requires less labeled data for continually training the model to achieve the outstanding performance. When the pre-trained model is trained on similar data in source domain, fine-tuning will generate improved performance on the target domain. (2) However, as a supervised method, fine-tuning requires the manually annotated ground truth on the target domain, which might be not available in some real-world applications. In addition, fine-tuning might suffer from model forgetting, where the model may miss some knowledge about the pre-trained task in source domain  when adapting to the new task in target domain. Furthermore, small target-domain dataset might result in the model overfitting during the fine-tuning.}

\subsubsection{Knowledge Distillation}
Knowledge distillation~\cite{hinton2015distilling,gou2021knowledge,wang2021knowledge,chen2021distilling,xie2022performance,beyer2022knowledge} is an advanced  technique in deep learning, which is also referred to as teacher-student learning, where a student neural network is trained on target domain to imitate the knowledge of a  teacher neural network trained on source domain. Knowledge distillation has been widely utilized in intelligent vehicle perception~\cite{kothandaraman2021domain,gao2022cross,hou2022point,yang2022cross,sautier2022image}. 

Knowledge distillation could be beneficial to model generalization, model compression, model transferability. It improves the model generalization so that the student network can generalize better on unseen examples, especially in scenarios with limited training data. It allows to compress a large teacher model into a smaller student model. It enables the knowledge transferability from the teacher model (source domain) to the student model (target domain) even with different deep learning architectures. The teacher network is typically trained on a large-scale dataset for the next knowledge transferability to the student network, however the large-scale dataset might be not available in the source domain of some intelligent vehicle perception tasks. Inspired by~\cite{lan2022instance}, the knowledge distillation methods could be roughly classified into the following two types.

\textit{Response Knowledge Distillation}: It focuses on the final output layer of the teacher model so as to teach a student model to mimic its predictions. The core concept is to use a loss function called the distillation loss, which measures the difference between the output activations of the student and teacher models. By minimizing this loss during training, the student model gradually improves its ability to generate predictions that closely resemble those of the teacher model. \cite{gao2022cross} proposes the cross-domain correlation distillation loss to transfer knowledge from daytime to nighttime domains, thereby improving nighttime semantic segmentation performance.

\textit{Intermediate Knowledge Distillation}: It focuses on aligning the intermediate representations of the teacher and student models. The intermediate layers learn to recognize and distinguish specific features in the data, and this knowledge distilled in teacher network can be leveraged to train the student model effectively. \cite{hou2022point} proposes an approach of transferring distilled knowledge from a larger source teacher model to a smaller target student network to conduct LiDAR semantic segmentation. Specifically, the intermediate Point-to-Voxel Knowledge Distillation approach is utilized to transfer latent knowledge from both point level and voxel level to complement sparse supervision signals.

\textcolor{black}{\textbf{In-depth Discussion:} (1) By the knowledge transfer from the teacher model to student model, knowledge distillation allows for model compression, improved generalization, regularization to prevent overfitting, and ensemble effect mimic. A smaller student model can be trained to approximate the performance of a larger computationally expensive teacher model, leading to the  model compression for the student model. By learning from the guidance provided by the teacher model, the student model will inherit the teacher's generalization abilities to unseen examples, ensuring better generalization of the student model. The guidance by the teacher model is able to act as a form of regularization that prevents overfitting of the student model. Knowledge distillation can also be combined with traditional model training to mimic the ensemble effect of multiple models. (2) However, the success of knowledge distillation relies on a well-trained high-performance  teacher model first. Training a larger teacher model can be computationally expensive, and the large-scale dataset might be not available in the source domain of some intelligent vehicle perception tasks. If the teacher model lacks sufficient recognition capacity (not well-optimized), then the knowledge transferred to the student model may be not strong enough. In some complex cases of intelligent vehicle perception tasks, knowledge distillation might struggle to improve the student model's performance when the data distribution difference between source and target domains is quite large.}

\subsection{Unsupervised TL}
In the intelligent vehicle perception, data labeling is a time-consuming and labor-intensive process in real-world scenarios. Generally, supervised algorithms struggle when there is a scarcity of labeled data in the source domains~\cite{niu2020decade,pan2010survey}. To overcome these challenges, Unsupervised Transfer Learning (TL) has emerged as a promising approach for addressing such specific cases in the  intelligent vehicle perception tasks. Unsupervised TL refers to a scenario where there is unlabeled target data besides labeled data available in source domain. Unsupervised TL approaches offer promising solutions to overcome the limitations of limited labeled data availability, enabling more efficient and effective perception in intelligent vehicles.

\subsubsection{Image-to-image Transfer}
Image-to-image transfer, also known as image-to-image translation, is a computer vision task that involves converting an input image to a different domain. It aims to establish a learned correspondence between two visual domains, where the input image originates from the source domain, while the desired output image that resembles the target domain. The goal is to generate a corresponding image with similar style of the target domain and simultaneously preserve the relevant characteristics and semantic contents of the input image. It has found extensive application in the field of autonomous driving as well as intelligent transportation systems, including semantic segmentation~\cite{murez2018image,pizzati2020domain}, lane recognition~\cite{hou2019learning,liu2021condlanenet}, data augmentation~\cite{zhang2022simbar,yang2020surfelgan}~\cite{mușat2021multi} and object detection~\cite{schutera2020night,li2021domain,li2022cross,shan2019pixel}.

Image-to-image transfer neural networks are commonly implemented using two different approaches: (1) 
\textit{Paired Image-to-Image Transfer} and (2)  \textit{Unpaired Image-to-Image Transfer}. The first approach utilizes generative adversarial networks trained on paired images~\cite{wang2018high}. This type of network learns a mapping that transforms an input image from its original domain to desired output domain~\cite{isola2017image}. The second approach addresses scenarios where unpaired images are used to establish a more general framework~\cite{zhu2017unpaired,park2020contrastive}, inspiring the unsupervised image-to-image translation  methods~\cite{liu2017unsupervised,baek2021rethinking}.


\textit{Paired Image-to-Image Transfer}: \cite{isola2017image} investigated the utilization of conditional Generative Adversarial Networks (GAN) namely pix2pix for paired image-to-image translation  tasks~\cite{hao2019learning}. The GAN with condition learns a generative model of data but with the added condition of an input image to produce a corresponding output image. This approach strives to produce plausible images in target domain. The adversarial loss is utilized to train a Generator Network which is updated using $l_1$ loss, which quantifies the disparity between the generated image as well as predicted output. By incorporating additional loss, the Generator Network  can produce plausible translations of the source images. Conversely, the Discriminator Network  is designed to perform generated  image classification. With the paired training data, these methods could translate the image of similar styles in different domains.

\textit{Unpaired Image-to-Image Transfer}: Cycle-consistency GAN (CycleGAN)~\cite{zhu2017unpaired} is a type of GAN model that enables image translation between unpaired  datasets~\cite{mușat2021multi,uricar2021let,shan2019pixel,liu2022multi}. The training process of a CycleGAN involves optimizing two generators and two discriminators simultaneously. One generator is responsible for learning the mapping function $G$ from domain $X$ to  $Y$, while the other generator $F$ learns the mapping from domain $Y$ to  $X$. Both  $G$  and $F$ are trained simultaneously, incorporating a cycle consistency loss that enforces the cycle consistency to ensure that $F(G(x)) \approx x$ and $G(F(y)) \approx y$. This loss combined with adversarial losses on domains $X$ and $Y$ yields objective for unpaired image-to-image translation. Unpaired Image-to-Image Transfer release the requirement of paired training data, which is more general in the real-world applications of intelligent vehicle perception. By incorporating adversarial losses on domains $X$ and $Y$, the objective for unpaired image-to-image translation is obtained. Unpaired Image-to-Image Transfer release the need for paired training data, making them more general in real-life applications of intelligent vehicle perception.

\textcolor{black}{\textbf{In-depth Discussion:} (1) Image-to-image transfer can be used as data augmentation, data distribution style transfer. Because of the limited data of source or target domain, image-to-image transfer methods could generate diverse fake examples as data augmentation to improve the robustness and generalization during the machine learning model training. In addition, image-to-image transfer enables the translation between source and target domains, to reduce the data distribution style difference in intelligent vehicle perception. (2) However, there are some limitations for the 
image-to-image transfer methods when applying to the real-world intelligent vehicle perception. When the transfer involves complex or ambiguous patterns in autonomous driving, image-to-image transfer models might produce translations with low realism or fidelity. For the \textit{Paired Image-to-Image Transfer} methods, in practical applications of intelligent vehicle perception, the requirement for paired training data poses disadvantages. For the \textit{Unpaired Image-to-Image Transfer} methods, they rely on task-specific and predefined similarity functions between inputs and outputs and do not consider the reliability and robustness of the translation frameworks, which might be disrupted by the perturbations added to input and targeted images. This issue is particularly crucial for autonomous driving.}

\subsubsection{Adversarial Learning}
Adversarial learning refers to a machine learning technique that involves training two neural networks in a competitive manner, which is initially introduced in the context of Generative Adversarial Networks (GAN) by~\cite{goodfellow2020generative} and also mentioned in Gradient Reversal Layer (GRL) framework~\cite{ganin2015unsupervised}, and provides a promising approach for generating target-similar samples at the pixel-level or target-similar representations at the feature-level by training robust deep neural networks. It has become popular for addressing transfer learning challenges by minimizing the domain discrepancy using adversarial objectives, such as fooling a domain discriminator/classifier. During training, the feature extractor and the domain discriminator are engaged in an adversarial game. The feature extractor tries to produce representations that confuse the domain discriminator, making it difficult for the discriminator to differentiate between the domains. Meanwhile, the objective of the domain discriminator is to correctly classify the samples into their respective domains. This adversarial process encourages  the learning of domain-invariant features by the feature extractor, thereby minimizing the differences between domains. By minimizing the domain disparities through adversarial learning, the model learns representations that capture the underlying domain-invariant information shared across domains. This approach helps to address TL challenges by effectively reducing the disparities between two different domains, improving the model's generalization capabilities across different domains. The adversarial learning based transfer learning methods for intelligent vehicle perception consists of two types: \textit{GRL based Methods} and \textit{GAN based Methods}. 

\textit{GRL based Methods}: Domain adaptation in different vehicle perception domains can be achieved through the addition of a Gradient Reversal Layer (GRL) to the deep learning architecture~\cite{xu2023bridging,li2023domain}. The mechanism of domain adversarial embedding involves using a discriminator with a GRL to differentiate between samples from two domains. The discriminator is a binary classifier, while the GRL can reverse the training gradient in the back propagation of feature extraction. Both the discriminator and the GRL work together to align the feature distributions across different domains. It is worth mentioning that the GRL only comes into effect during the backpropagation phase and does not affect the forward propagation process~\cite{ganin2015unsupervised}. Let us give a detailed example for better understanding. \cite{li2023domain} introduces a new framework for domain adaptive object detection in autonomous driving during challenging foggy weather. The approach addresses the domain gap between clear and foggy weather in vehicle   driving by incorporating image-level and object-level adaptation techniques, which aim to minimize differences in object appearance and image style. Additionally, a novel Adversarial Gradient Reversal Layer (AdvGRL) has been proposed to enable adversarial mining for difficult examples along with domain adaptation.

\textit{GAN based Methods}: GAN~\cite{goodfellow2020generative,song2020multi} is a popular deep learning framework that can be used to teach a model to capture the distribution patterns present within the training data, enabling the generation of new data from that same distribution. A GAN consists of two separate models, namely the generator $G$ and the discriminator $D$. The generator $G$'s job is to create ``fake" images that resemble the training images so as to confuse the discriminator $D$. The applications of GAN in  autonomous driving have  been recently explored owing to its remarkable progress in generating realistic images. Specifically, GAN has been leveraged to generate image or subspace feature undistinguished by domain classifier based discriminator, for example, GAN could generate aligned/similar features between clear weather and foggy weather~\cite{li2023domain,li2022stepwise}, between synthetic game data and real-world data~\cite{biasetton2019unsupervised,zhang2021target}, between daytime data and nighttime data~\cite{wang2022sfnet,li2022stepwise}. Let us give a detailed example for better understanding. \cite{hoffman2018cycada} proposes a domain adaptation model which combines generative image space alignment,  latent feature space alignment, and the vehicle perception task. By considering the vehicle perception task (semantic segmentation of urban driving scenes), the image-level features, latent features, and the task-related semantic features are aligned across different domains by an adversarial learning via a GAN-based framework.  

\textcolor{black}{\textbf{In-depth Discussion:} (1) The \textit{GRL based Methods} rely on minimizing the domain distribution discrepancy through a gradient reversal in the back propagation of feature extraction to confuse the domain discriminator. In contrast, the \textit{GAN  based Methods} focus on training the Generator Network and Discriminator Network alternately using a Min-Max adversarial loss function, with the goal of acquiring domain-invariant features. In these two ways, the feature  distribution between source and target domains can be aligned to reduce domain gap for transfer learning. (2) However, training the adversarial learning model is sometimes difficult. For the \textit{GRL based Methods}, the simple gradient reversal might be not powerful enough to well minimize the domain distribution discrepancy in complex driving scenarios, leading to the local optimal solution. For the \textit{GAN based Methods}, finding the optimized balance between the generator $G$ and discriminator $D$ can be difficult and the convergence may not always be guaranteed when training the GAN model, due to its hyperparameter sensitivity, data quality, and data diversity.}

\subsubsection{Feature Alignment}
 
To minimize the domain distribution discrepancy, the objective of feature alignment in transfer learning is to discover an aligned feature representation from multiple domains. Typically, the feature distribution difference between different domains can be defined as loss functions during the deep neural network training, so minimizing the loss functions of the feature distribution difference across multiple domains will reduce the domain gap. 

Feature alignment-based transfer learning can be classified into two main categories: \textit{Subspace Feature Alignment}, and \textit{Attention-guided Feature Alignment}.

\textit{Subspace Feature Alignment}: By projecting the features from different domains to a lower-dimensional subspace, several metrics to describe the distance of feature distribution across source and target domains can be defined as the loss functions in the deep learning framework. Minimizing these metric distances (loss functions) will align the features of different domains in the subspace. The widely used metric to describe the feature distribution distances are Principal Component Analysis (PCA) projected subspace feature distance~\cite{song2019domain}, Maximum Mean Discrepancy (MMD)~\cite{borgwardt2006integrating}, Kullback–Leibler Divergence ~\cite{zhang2018deep}, Gram Matrix~\cite{guo2019degraded}, Multi-Kernel MMD~\cite{gretton2012optimal,long2015learning}, Joint MMD~\cite{long2017deep}, Wasserstein distance~\cite{arjovsky2017wasserstein}, \textit{etc}. For example, let us take a close look at the definition of the MMD metric, which is formulated as  

\begin{equation}
MMD(\mathcal{X}_s, \mathcal{X}_t) =  \lVert \frac{1}{n_s} \sum_{i=1}^{n_s} k(\mathbf{x}_i^s) - \frac{1}{n_t} \sum_{j=1}^{n_t} k(\mathbf{x}_j^t) \rVert_{H},   
\end{equation}
where $\mathcal{X}_s$ and $\mathcal{X}_t$ denote the sets of samples obtained from the source and target domains,  $\mathbf{x}_i^s$ and $\mathbf{x}_j^t$ are individual samples from the respective domains, and $n_s$ and $n_t$ denote the sample sizes of the source and target domains respectively, $k$ denotes the kernel functions, and $H$ indicates the Reproducing Kernel Hilbert Space (RKHS).    

\textit{Attention-guided Feature Alignment}: Taking inspiration from the attention mechanism~\cite{zhou2016learning,vaswani2017attention}, the most informative components of specific importance can be focused for the intelligent vehicle perception. The deep learning frameworks can first extract the attention maps, then the distance of attention maps between two domains can be defined as loss function to be minimized during the neural network training~\cite{zhou2020multi,zagoruyko2016paying}. 
By employing this approach, it becomes possible to align the feature distribution across both the source and target domains via the attention map consistency constraint. 
For example, in~\cite{cho2023itkd}, the relation-aware knowledge captured by multiple detection heads can be transferred using a specially designed  attention head loss for the improved LiDAR-based 3D object detection in the context of autonomous driving.  

\textcolor{black}{\textbf{In-depth Discussion:} (1) The \textit{Subspace Feature Alignment} methods focus on aligning the feature distribution in the lower-dimensional subspace representation by using different metrics of distribution distances. The \textit{Attention-guided Feature Alignment} methods use the attention mechanism to extract the attention maps first and then enforce the attention maps from multiple domains to be the same. In these two ways, feature alignment facilitates the model to adapt its knowledge learned from the source domain to the target domain, so the model can generalize better to the target domain. (2) However, several important settings in the feature alignment methods are still open questions. Let us give some examples. How to discover the most representative feature subspaces or attention maps does not have a widely-accepted common sense in the diverse deep neural network architectures. The feature distribution distance metrics (like MMD) are good but may still  ignore some important domain information. How to balance the loss weights of the feature alignment and the original task related loss for intelligent vehicle perception is still challenging during the model training.} 

\subsubsection{Self-learning}
Autonomous vehicles continuously collect unlabeled data during their operation, creating an opportunity for self-learning~\cite{liu2021source,zhang2021transfer,kumar2021syndistnet,luo2021self,ziegler2022self}, which offers a promising approach to reduce the reliance on labeled data and enhance model flexibility. Given the absence of labeled data in target domain using Unsupervised TL, the self-learning methods use the additional cues to evaluate the neural network prediction in an unsupervised setting, so some prediction results with high confidence are used as the pseudo-labels in the further training or testing.

The following shows some representative examples of self-learning methods for the Unsupervised TL based intelligent vehicle perception. The entropy based uncertainty can be used to define the hardness of a specific training sample so as to implement an easy-to-hard curriculum learning for semantic segmentation~\cite{pan2020unsupervised}. \cite{wang2021domain} utilizes self-supervised learning to enhance the semantic segmentation performance by using depth estimation as guidance to overcome the domain gap between the source and target domains. They explicitly capture the correlation between task features and use target depth estimation to enhance target semantic predictions. The adaptation difficulty, as inferred from depth information, is subsequently utilized to enhance the quality of pseudo-labels for target semantic segmentation. \cite{shin2022mm} proposes a multi-modal extension of test-time adaptation in the context of 3D semantic segmentation. To improve the unstable performance of models at test time, they design both intra-modal and inter-modal modules together to acquire more dependable self-learning signals of pseudo-labels. \cite{zhang2021transfer} utilizes the multiple classifiers with attention heads to evaluate the uncertainty associated with the pseudo-labels. The panoramic pseudo-labels with high confidences are then used to improve the panoramic semantic segmentation prediction in an  iterative fashion.

\textcolor{black}{\textbf{In-depth Discussion:} (1) Self-learning bridges the gap between supervised and unsupervised learning, which combines the advantages of the both methods by using  labeled data for training and unlabeled data for further refinement. By leveraging self-learning in autonomous driving, the need for extensive manual annotation of data is reduced, enabling more cost-effective and efficient training of models. The iterative process of utilizing high-confidence identified samples and generating pseudo-labels facilitates promising methods using unlabeled data in target domain. (2) However, the robustness and convergence of the self-learning methods is still an open question for the reliable intelligent vehicle perception. If the initial model is not sufficiently confident and makes mistake  predictions on unlabeled data, such errors might be propagated through iterations, leading to poor performance or hard convergence.}

\subsection{Weakly-and-semi Supervised TL} 

Although impressive results have been achieved by unsupervised TL methods, the domain gap cannot be completely eliminated due to the lack of supervision on the target domain. There is still a relative  performance gap compared with supervised TL methods. Another way in addressing the domain gap is by using the weakly-and-semi supervised learning method that utilizes both weakly labeled and some labeled/unlabeled data in target domain. Based on the available supervision, the weakly-and-semi supervised transfer learning methods could be roughly classified into two types: \textit{Weakly-Supervised TL}: There are only weakly supervised labels in the target domain. \textit{Semi-Supervised TL}: There are only semi-supervised labels in the target domain, including some labeled data and the remaining unlabeled data on target domain.

\textit{Weakly-Supervised TL}: Theories of weakly supervised learning have been applied in autonomous driving~\cite{barnes2017find,gojcic2021weakly}, such as object detection, semantic segmentation, and instance segmentation. The transfer learning techniques can be applied simultaneously with the weakly supervised learning. For example, when an instance-level task only has image-level annotations in target domain but with instance-level annotations in source domain, the pseudo annotations can be  predicted~\cite{inoue2018cross} for the object detection task. Given a source domain (synthetic data) with pixel/object- level labels, a target domain (real-world scenes) might only have object-level labels, where the pixel-level and object-level domain classifiers can be used in transfer learning to learn domain-invariant features for the semantic segmentation task in driving scenes~\cite{wang2019weakly}.

\textit{Semi-Supervised TL}: There are three types of training data (labeled source data, labeled target data, and unlabeled target data) in the semi-supervised TL setting~\cite{wang2020alleviating,chen2021semi,wang2023ssda3d}. The key point for improving semi-supervised TL is to effectively use available unlabeled data from target domain and limited labeled data from different domains. For example, \cite{wang2020alleviating} aligns feature distribution across two domains by introducing an extra semantic-level adaptation module, which leverages a few labeled images from the target domain to supervise the segmentation and feature adaptation tasks. Other works focus on generating pseudo labels for unlabeled target data by using labeled source data and labeled target data. For example, \cite{wang2023ssda3d} solves this problem by two-stage learning that includes inter-domain adaptation stage and intra-domain generalization stage. While \cite{chen2021semi} uses the domain-mixed teacher models and knowledge distillation to train a good student model, then the good student model will generate pseudo labels for the next round of teacher model training.  

\textcolor{black}{\textbf{In-depth Discussion:} (1) By involving some supervisions in the target domain, the weakly-and-semi supervised learning methods could achieve a better performance than the unsupervised TL methods. \textit{Weakly-Supervised TL} can be more data-efficient as it leverages weakly labeled data which is often easier and cheaper to obtain, reducing the need for extensive manual labeling. \textit{Semi-Supervised TL} can improve the model's generalization to unseen data and reduce overfitting on the limited labeled data. Semi-supervised methods augment the labeled dataset with unlabeled data, providing the model with additional training examples and increasing data diversity. Similar to weakly supervised methods, semi-supervised methods can be valuable for domain adaptation tasks, when limited labeled data is available in the target domain. (2) While various methods have been proposed for weakly-and-semi supervised transfer learning, how to leverage the unlabeled target data with the help of available labeled data under different complex situations is still challenging. The performance of weakly-and-semi supervised transfer learning methods still perform worse than the supervised transfer learning methods, which indicates that some local optimal solutions are achieved only. How to make a further improvement to overcome the algorithm performance boundary is an open question now. Maybe some prior knowledge of human driving can be leveraged to advance the algorithm performance.}

\subsection{Domain Generalization}

Domain Generalization (DG) for intelligent vehicle perception offers a solution to the challenge of enhancing the resilience of deep neural networks against arbitrary unseen driving scenes~\cite{zhou2022domain}. Unlike Domain Adaptation (DA), DG methods typically focus on learning a shared representation across multiple source domains. This approach aims to enhance the model ability to generalize across various domains, enabling it to perform well in an unknown target domain of driving. Nevertheless, the collection of multi-domain datasets is a laborious and costly endeavor, and the efficacy of DG methods is significantly influenced by the quantity of source datasets~\cite{wang2022generalizing}.

The concept of domain generalization (DG) has emerged as a solution to address the lack of target data in domain gap~\cite{blanchard2011generalizing}~\cite{wang2022generalizing}. The primary distinction between DA and DG lies in the fact that DG does not require access to the target domain during the training phase. DG aims to develop a model by using data from one or multiple related but distinct source domains to generate any out-of-distribution target domain data~\cite{shen2021towards}. The existing methods for DG can be divided into two main groups according to the number of source domains: \textit{Multi-source DG} and \textit{Single-source DG}. 

\textit{Multi-source DG}: Its primary motivation is to utilize data from multiple sources to learn representations that are invariant to different marginal distributions~\cite{wilson2020survey}~\cite{luo2022towards}~\cite{zhao2022style}. Due to the absence of target data, it is challenging for a model trained on a single source to achieve generalization effectively. By leveraging multiple domains, a model can discover stable patterns across the source domains, leading to better generalization results on unseen domains. The underlying concept behind this category is to minimize the difference between the representations of various source domains, thus learn domain-invariant representations~\cite{yue2019domain,hu2022causal,xu2022dirl,li2022cross,choi2021robustnet,lin2021domain,acuna2021towards}.

\textit{Single-source DG}: It assumes that the training data is homogeneous, which is sampled from a single domain~\cite{qiao2020learning,wang2021learning}. Single-source DG methods revolve around data augmentation, and they aim to create samples that are out of the domain and utilize them to train the network in conjunction with the source samples, enhancing the generalization capability~\cite{li2023intra,lehner20223d,hu2022causal,khosravian2021generalizing,chuah2022itsa,sanchez2022domain,zhang2020exploring,wu2022single}. Although single-source DG methods are not robust as multi-source domain method due to the limited information from source domain, they do not rely on domain identity labels for learning, which makes them applicable to both single-source and multi-source scenarios.

\textcolor{black}{\textbf{In-depth Discussion:} (1) By learning domain-invariant features plus data augmentation or random feature generalization, domain generalization aims to enhance the model's ability to be generalized to new unseen data. This can be beneficial for intelligent vehicle perception because many complex driving scenarios in the real world are not seen before. By incorporating \textit{Multi-source DG} and \textit{Single-source DG}, they can reduce the data bias in the existing training data, so the perception model can be more robust to new unseen driving scenarios. It is impossible to collect all the driving data in the real world, so domain generalization potentially saves time and labor cost with improved model robustness. (2) However, training a perception model with the domain generalization capability is more complex than traditional domain-specific transfer learning methods, as it requires handling the generalization of domain-invariant features for either single source or multiple sources. When training on multiple source domains, the model might suffer from the data imbalance across different domains, leading to biased learning towards some dominant domains. This issue is similar for the class imbalance problem in a single source domain, posing challenges to the domain generalization methods.}

\begin{table}[h]
\centering
\footnotesize

\caption{\textcolor{black}{Uniqueness of Deep Transfer Learning (TL) Methods for Intelligent Vehicle Perception.}}

\begin{adjustbox}{width=1\textwidth}
\begin{tabular}{c|c|c}
\hline
\hline
\textbf{Types} & \textbf{Methodologies} & \textbf{Uniqueness} \\
\hline
\multirow{2}{*}{Supervised TL} & Fine-tuning & Continuous learning from pre-trained model\\
\cline{2-3}
& Knowledge distillation & Knowledge transfer from teacher to student model\\
\hline
\multirow{4}{*}{Unsupervised TL} & Image-to-image transfer & Pixel-level mapping or translation \\
\cline{2-3}
& Adversarial learning &  Training two models adversarially\\
\cline{2-3}
& Feature alignment & Aligning features in different domains\\
\cline{2-3}
& Self-learning & Learning with pseudo-labels for refinement\\
\hline
\multirow{2}{*}{Weakly-and-semi Supervised TL} & Weakly-Supervised TL & Learning with weakly-labeled data \\
\cline{2-3}
& Semi-Supervised TL & Learning with partially-labeled data\\
\hline
\multirow{2}{*}{Domain Generalization} & Multi-source DG & Generalization to unseen data  by  multiple source domains\\
\cline{2-3}
& Single-source DG & Generalization to unseen data  by single source domain \\
\hline
\hline
    \end{tabular}
    \end{adjustbox}
\label{tab:uniqueness_method}
\end{table}

\subsection{\textcolor{black}{Uniqueness of Methodology}}
\textcolor{black}{The Table~\ref{tab:uniqueness_method} summarizes the uniqueness of the above classified deep transfer learning methods for intelligent vehicle perception, including the methods of Supervise TL, Unsupervised TL, Weakly-and-semi Supervised TL, and Domain Generalization.}

\section{\red{Challenges and Future Research}} \label{sec:Cha}
This section outlines the main challenges of the deep transfer learning for the current intelligent vehicle perception \textcolor{black}{and the related future research directions}.

\begin{itemize}
    \item \textbf{Sensor Robustness}: The current  camera and LiDAR sensors are not robust enough in the extreme driving scenarios, like diverse weather, dark illumination, various environments. In addition, for the V2V cooperative perception, the V2V communication sensors might have the issues of lossy communication~\cite{li2023learning} due to the fast speed, obstacles, \textit{etc}~\cite{schlager2022automotive}. \textit{Future Research:} More future research can be focused on improving the sensor robustness, for example, the camera and LiDAR sensors in  diverse weather, dark illumination, various environments, and the communication sensors in the V2V system~\cite{tahir2021deployment}.  \textcolor{black}{For example, more advanced sensors with robust lens coatings, self-cleaning mechanisms, better LiDAR reflection can be studied to compensate for distortions in  adverse weather. More advanced V2V communication systems with less communication delay can be investigated, anticipating the future growth of autonomous connected vehicles.}

    \item \textbf{Methodology Limitation}: The current unsupervised transfer learning methods are worse than the supervised transfer learning methods with a relative performance insufficiency. In addition, how to fully utilize the knowledge of the source domain and the human prior cognition and experience is still a question to be answered. How to effectively use  the weakly and partially labeled data is still a open question. \textit{Future Research}: Researchers could make efforts to develop more advanced deep transfer learning methods in the future, for example, largely reducing the performance disparity between unsupervised and supervised approaches, \textcolor{black}{incorporating the  Vehicle-to-Everything (V2X) techniques to communicate with connected vehicles and smart infrastructures to overcome the occlusion challenges}, involving the Large Language Models, like ChatGPT~\cite{gao2023chat}, to better simulate the human cognition and knowledge so as to guide the transfer learning methods, accurately self-learning the unlabeled data, effectively and efficiently using the weakly and partially supervised data.   

    \item \textbf{Realism of Synthetic Data}: By eliminating the need for manual annotation, the synthetic data generated by computer game engines is quite helpful to improve the training data size, but it still has significant differences with the real-world data in styles, lighting conditions, viewpoints, and vehicle behaviors, \textit{etc}. \textit{Future Research}: The realism of the synthetic data can be improved by more advanced computer game engines in the future. The customized synthetic data can be better simulated via a digital twin simulation system~\cite{wang2023new}. \textcolor{black}{For example, researchers can engage in the development of dynamic and interactive virtual environments within gaming engines. Through the simulation of authentic variations in illuminating, weather conditions, and object interactions, synthetic data will be able to emulate the complexities of real-world scenarios. This process could create more robust and transferable machine learning models. Moreover, constructing digital twin models based on real-world data offers the opportunity to generate highly personalized and precise synthetic data.}
    
    \item \textbf{Scarcity of Annotated Benchmarks in Complex Scenarios}: There are infinite complex scenarios in the real-world driving, but the current benchmark datasets in the complex driving scenarios are still limited. For example, the Foggy Cityscapes dataset~\cite{sakaridis2018semantic} only has 2,975 training images during the foggy weather, whose small size poses a clear hurdle for the accurate perception of the intelligent vehicle in the foggy weather. \textit{Future Research}: We expect that more high-quality benchmark datasets in complex driving scenarios could be collected and publicized in the future. We also encourage more advanced physical models to simulate the benchmark data (Camera, LiDAR) in complex driving scenarios in the future, such as simulating the fog, rain, snow, lighting changes, \textit{etc}. \textcolor{black}{For example, future research may focus on collecting large-scale real-world benchmark datasets (Camera, LiDAR) that cover a wide range of complex driving scenarios with  challenging urban or highway environments, dense traffic, pedestrian crossings, various traffic patterns and different road geometries.}

    \item \textbf{International Standards for Hardware Sensors}: The hardware sensors might be provided  from multiple companies of different countries, but there are no unified international standards for the hardware sensors for intelligent vehicle perception. For example, the different hardware sensor types and settings will enlarge the domain gap in different environments. \textit{Future Research}: We hope that the multiple companies of different countries can collaborate together to promote the international standards for hardware sensors in the future, including the types, settings, parameters of the hardware sensors in different driving environments~\cite{schlager2022contaminations,masmoudi2021reinforcement}. \textcolor{black}{Compatible hardware sensor standards would enable the exchange and sharing of data across different intelligent  vehicle platforms, allowing for improved decision-making and safety on the roads.}

    \item \textbf{International Standards for Software Packages}: The software package might be provided  from multiple companies of different countries as well, but there are no unified international standards for the software packes for intelligent vehicle perception. For example, sharing the features of models trained in different epochs, \textit{e.g.}, from different companies, will result in the performance drop in V2V cooperative perception~\cite{xu2023bridging}. \textit{Future Research}: The multiple companies of different countries are expected to collaborate together to promote the international standards for software packages in the future, including the deep learning model architectures and frameworks, hyper parameters, privacy and safety preservation, \textit{etc}. \textcolor{black}{For example, the accepted software package standards can reduce the problems in the V2V data sharing of the connected intelligent vehicles among different companies with guidelines.}
    
\end{itemize}

\section{Conclusion}\label{sec:Con}
In this survey paper, we presented a comprehensive review of deep transfer learning for intelligent vehicle perception. We reviewed the perception tasks and the related benchmark datasets and then divided the domain distribution discrepancy of the intelligent vehicle perception in the real world    into sensor, data, and model differences. Then, we provided clearly classified and summarized definition and description of numerous representative deep transfer learning  approaches and related works in intelligent vehicle perception. Through our intensive analysis and review, we have identified several potential challenges and directions for future research. Overall, this survey paper aims to make contributions to introduce and explain the deep transfer learning techniques for intelligent vehicle perception, offering invaluable insights and directions for the future research.  

\section{CRediT authorship contribution}\label{sec:CRe}
Xinyu Liu: Conceptualization, Methodology, Original draft preparation. Jinlong Li: Methodology, Original draft preparation. Jin Ma: Methodology, Investigation. Huiming Sun: Methodology, Investigation. Zhigang Xu: Review \& editing. Tianyun Zhang: Review \& editing. Hongkai Yu: Methodology guidance, Supervision, Review \& editing. 


\section{Declaration of Competing Interest}\label{sec:Dec}
The authors declare that they have no known competing financial interests or personal relationships that could have appeared to influence the work reported in this paper.

\section{Acknowledgement}\label{sec:ack}
This work was supported by NSF 2215388 and CSU FRD grants. 
 
\bibliography{xinyu_bib}

\end{document}